\pdfoutput=1
\documentclass[11pt]{article}

\usepackage[preprint]{acl}

\usepackage{times}
\usepackage{latexsym}
\usepackage{amsmath}
\usepackage{amssymb}
\usepackage{graphicx}
\usepackage{subcaption}
\usepackage{booktabs}
\usepackage[most]{tcolorbox}
\usepackage{enumitem}

\usepackage[T1]{fontenc}
\usepackage[utf8]{inputenc}

\usepackage{microtype}

\usepackage{inconsolata}

\title{DiffLoRA: Differential Low-Rank Adapters for Large Language Models}

\author{\textbf{Alexandre Misrahi}\textsuperscript{1}\thanks{Work done during internship at NAVER LABS Europe}\quad
  \textbf{Nadezhda Chirkova}\textsuperscript{2}\quad
  \textbf{Maxime Louis}\textsuperscript{2}\quad
  \textbf{Vassilina Nikoulina}\textsuperscript{2}\\
  $^1$ EPFL\quad$^2$NAVER LABS Europe\\
  \texttt{alexandre.misrahi@epfl.ch, vassilina.nikoulina@naverlabs.com}
}

\begin{document}
\maketitle
\begin{abstract}
Differential Transformer has recently been proposed to improve performance in Transformer models by canceling out noise through a denoiser attention mechanism. In this work, we introduce DiffLoRA, a parameter-efficient adaptation of the differential attention mechanism, with low-rank adapters on both positive and negative attention terms. This approach retains the efficiency of LoRA while aiming to benefit from the performance gains of differential attention. We evaluate DiffLoRA across a broad range of NLP tasks, including general benchmarks, many-shot in-context learning, RAG, and long-context tests. We observe that, although DiffLoRA falls short of other parameter-efficient fine-tuning methods in most evaluation tasks, it shows interesting results in certain domains (+11 pts on LoRA for HumanEval). We analyze the attention patterns post-finetuning to identify the reasons for this behavior. 
\end{abstract}

\section{Introduction}

Large language models (LLMs) have achieved remarkable success across diverse NLP tasks, but adapting these massive models to new domains or tasks remains challenging and costly. Full fine-tuning of an LLM for each application is often infeasible due to the large number of parameters. This drives the need for efficient and robust LLM adaptation techniques that can customize model behavior for a variety of tasks. Multiple parameter-efficient fine-tuning methods have emerged to address this challenge, the most prominent approach being LoRA \citep{hu2021loralowrankadaptationlarge}, which injects small trainable weight matrices into a pre-trained model instead of updating all weights. 

In parallel, recent architectural innovations like the Differential Transformer  \citep{ye2024differentialtransformer} have tackled the well-known issue of attention sinks \citep{xiao2024efficientstreaminglanguagemodels}. Differential Transformer introduces a differential attention mechanism (DiffAttn) that amplifies attention to important context while canceling out noise. This strategy demonstrated significant performance improvement in context-critical tasks such as Retrieval-Augmented Generation (RAG) or In-Context Learning (ICL), as well as remarkable domain robustness. However, a current limitation of this method is that it requires to train a model from scratch. 

In this work, we explore DiffLoRA\footnote{We release our code at \url{https://github.com/alexmsrh/difflora}}, a technique that integrates LoRA and DiffAttn to adapt pre-trained LLMs using low-rank adapters. As in \citep{grattafiori2024llama3herdmodels}, LoRA adapters are incorporated at each layer of the model, enabling it to learn the denoising weights associated with DiffAttn. The goal of DiffLoRA is to adapt a pre-trained model in a parameter-efficient manner, while aiming to match the performance improvements demonstrated by the Differential Transformer, and potentially outperform corresponding baselines in context-heavy tasks.

\section{DiffLoRA}

In this section, we provide a formal description for our method. DiffLoRA uses a similar attention function as Differential Transformer: 
\[
\text{DiffAttn}(X)=\left(\mathrm{sm}\left(\frac{Q_1K^T_1}{\sqrt{d}}\right)-\lambda\cdot\mathrm{sm}\left(\frac{Q_2K^T_2}{\sqrt{d}}\right)\right)V
\]
where $d$ is the model's hidden size used in transformer layers, and $\mathrm{sm}$ is the softmax function. In the setting of a pre-trained LLM, we obtain $Q_1$, $K_1$ from pre-trained $W_{Q_1}$, $W_{K_1}$ and aim to parameter-efficiently train the denoising terms $W_{Q_2}$, $W_{K_2}$ such that computation time and resources are similar to LoRA. To do so, we train low-rank adapters \citep{hu2021loralowrankadaptationlarge} $B_{Q_2},B_{K_2}\in\mathbb{R}^{N\times r}, A_{Q_2},A_{K_2}\in\mathbb{R}^{r\times d}$ such that 
\begin{align}\notag
    Q_2 = X(B_{Q_2}A_{Q_2})\\\notag
    K_2 = X(B_{K_2}A_{K_2})
\end{align}
We also add adapters on the positive term to increase the expressiveness of the adapters:
\begin{align}\notag
    Q_1 = X(W_{Q_1} + B_{Q_1}A_{Q_1})\\\notag
    K_1 = X(W_{K_1} + B_{K_1}A_{K_1})
\end{align}
where the new weights $ \{{A,B}\}_{\{{Q,K}\}_{\{1,2\}}} $ are initialized and trained as in LoRA. This is done for the attention mechanism at each hidden layer of the model.

\section{Experiments}
We train all models with a single epoch on Tulu-2 \cite{ivison2023camels}\footnote{\href{https://huggingface.co/datasets/allenai/tulu-v2-sft-mixture}{huggingface.co/datasets/allenai/tulu-v2-sft-mixture}} instruction tuning dataset. We perform an additional experiment with Tulu-3 \cite{lambert2024tulu3} to assess the impact of larger training dataset sizes. We rely on the open-instruct\footnote{\href{https://github.com/allenai/open-instruct}{github.com/allenai/open-instruct}} framework for finetuning.  The hyperparameters used for training are given in Appendix Table \ref{tab:hyperparams}. We proceed by describing the evaluation settings.

\subsection{General Evaluation}
We first evaluate the performances in terms of core LLM abilities, to investigate whether the fine-tuned model preserves its initial capabilities. We select a subset of datasets representing different types of knowledge encoded into LLMs: (1) Knowledge recall (TruthfulQA, PopQA, ARC-challenge), (2) reasoning (DROP, BBH), (3) math (GSM8k), (4) coding (HumanEval). We use the OLMES framework\footnote{\href{https://github.com/allenai/olmes}{github.com/allenai/olmes}} designed for evaluation reproducibility in LLMs. We rely on predefined evaluation settings/metrics for the above mentioned tasks. 

\subsection{Context-Sensitive Evaluation}
For In-Context Learning (ICL) and Needle-in-the-Haystack (NIH) tasks we rely on evaluation scheme from HELMET \citep{yen2025helmet}.

\paragraph{In-Context-Learning.}
 The TREC tasks \cite{li-roth-2002-learning} consist in classifying question type among 6 and 50 labels, respectively. The Clinic150 task \citep{larson-etal-2019-evaluation} and the Banking77 task \citep{casanueva-etal-2020-efficient} consist in classifying question intent among 151 and 77 classes, respectively. The task aims to evaluate the capability of DiffLoRA models to adapt to new tasks in a zero-shot fashion. As in \citep{yen2025helmet}, we report accuracy on the test sets. 

\paragraph{Needle-in-Haystack.}
We evaluate Needle-in-Haystack (NIH) performance across several settings: Multi-Key (MK) tasks the model to retrieve the correct key in the context with multiple noisy ones, and Multi-Value (MV) tasks the model to retrieve all the values associated with a certain key. 

\paragraph{RAG-QA}
We also evaluate our models in RAG Q\&A tasks, to assess their capability to exploit context and generate sound text. We follow RAG settings proposed in \cite{rau2024bergenbenchmarkinglibraryretrievalaugmented}, using BERGEN\footnote{\href{https://github.com/naver/bergen}{github.com/naver/bergen}} framework.  More details on RAG settings are available in the Appendix.  
We evaluate on both general QA benchmarks such as KILT-NQ \citep{petroni2021kiltbenchmarkknowledgeintensive} and PopQA \citep{mallen2023trustlanguagemodelsinvestigating}, as well as more specific domain benchmarks such as biomedical (BioASQ \citep{Nentidis_2023}), tech-support (TechQA \citep{castelli2019techqadataset}) and finance (\href{https://sites.google.com/view/fiqa/}{FiQA}). This evaluates the ability of DiffLoRA models to effectively answer questions by using context retrieved from a datastore.

\begin{figure*}[ht!]
    \centering
    \includegraphics[width=0.99\textwidth]{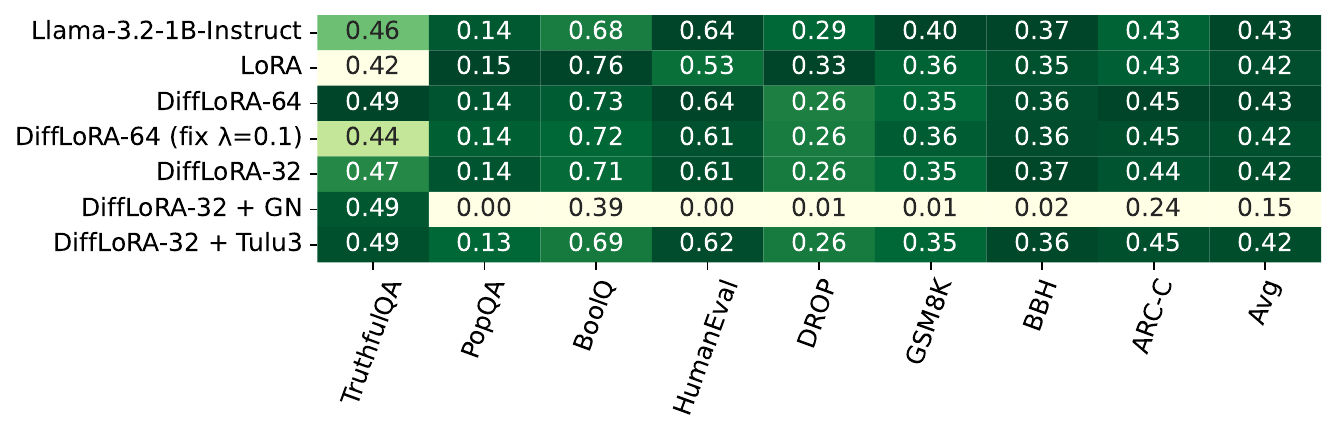}
    \caption{Evaluation of general LLM capabilities before and after finetuning. DiffLoRA-32: both right and left term of diff attention contain learnable parameters, DiffLoRA-64: only right term is learnable}
    \label{fig:main_res}
\end{figure*}

\subsection{Baselines}

We use Llama-3.2-1B-Instruct model\footnote{\href{https://huggingface.co/meta-llama/Llama-3.2-1B-Instruct}{meta-llama/Llama-3.2-1B-Instruct}} as starting point of our experiments. We compare the performance of DiffLoRA to this model to assess the impact of introducing denoiser adapters. In order to decouple the effect of finetuning from the effect of DiffAttn, we also perform LoRA finetuning on the same tuning datasets. We set Full LoRA rank in a way to match the number of trainable parameters of DiffLoRA models (more details at Appendix Tab. \ref{tab:hyperparams}).

\subsection{DiffLoRA Variants}

We hypothesize that some adaptation might be necessary in the positive term of attention in order to better adapt to the introduction of the negative side. To ensure that it is comparable with LoRA and the full denoiser setting in terms of number of parameters, we set the adapter rank for this variant to $r/2$, where $r$ is the rank for the setting with adapters only on the negative term. In our experiments we set $r=64$.  In \cite{ye2024differentialtransformer} an extra normalization (Group Norm) applied to each head independently to stabilize scale across heads, and a corresponding scaling factor of $(1-\lambda_{init})$ to stabilize this normalization in the gradient. We therefore add a model with Group Norm (GN) for comparison (more details in Appendix \ref{tab:hyperparams}). Following \citep{ye2024differentialtransformer}, in our experiments we learn the parameter $\lambda$, however we generally observe more stable results by freezing $\lambda$ to a small fixed value ($0.1$).

\begin{figure}
    \centering
    \includegraphics[width=0.99\linewidth]{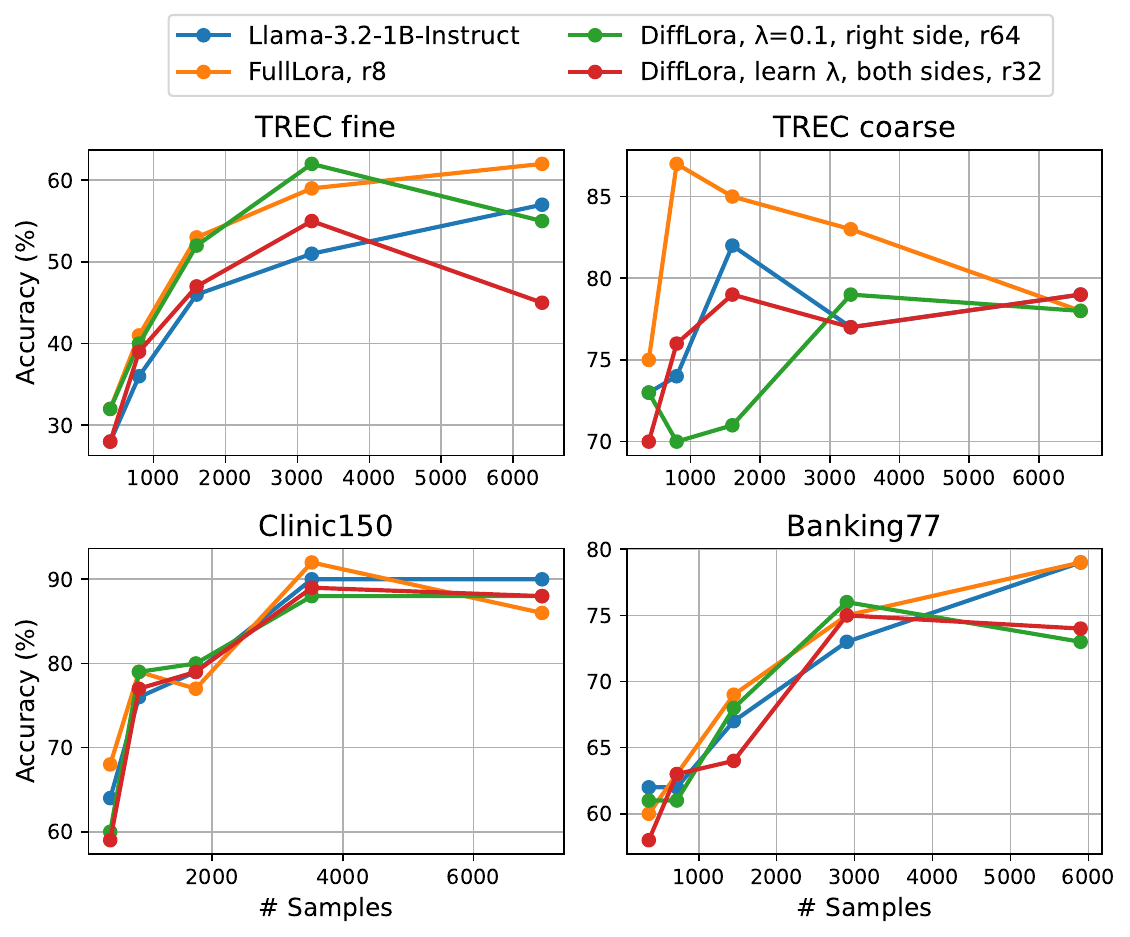}
    \caption{Evaluation on Many-shot In-Context Learning}
    \label{fig:many-shot-icl}
\end{figure}

\section{Results}

\begin{figure}
    \centering
    \includegraphics[width=0.99\linewidth]{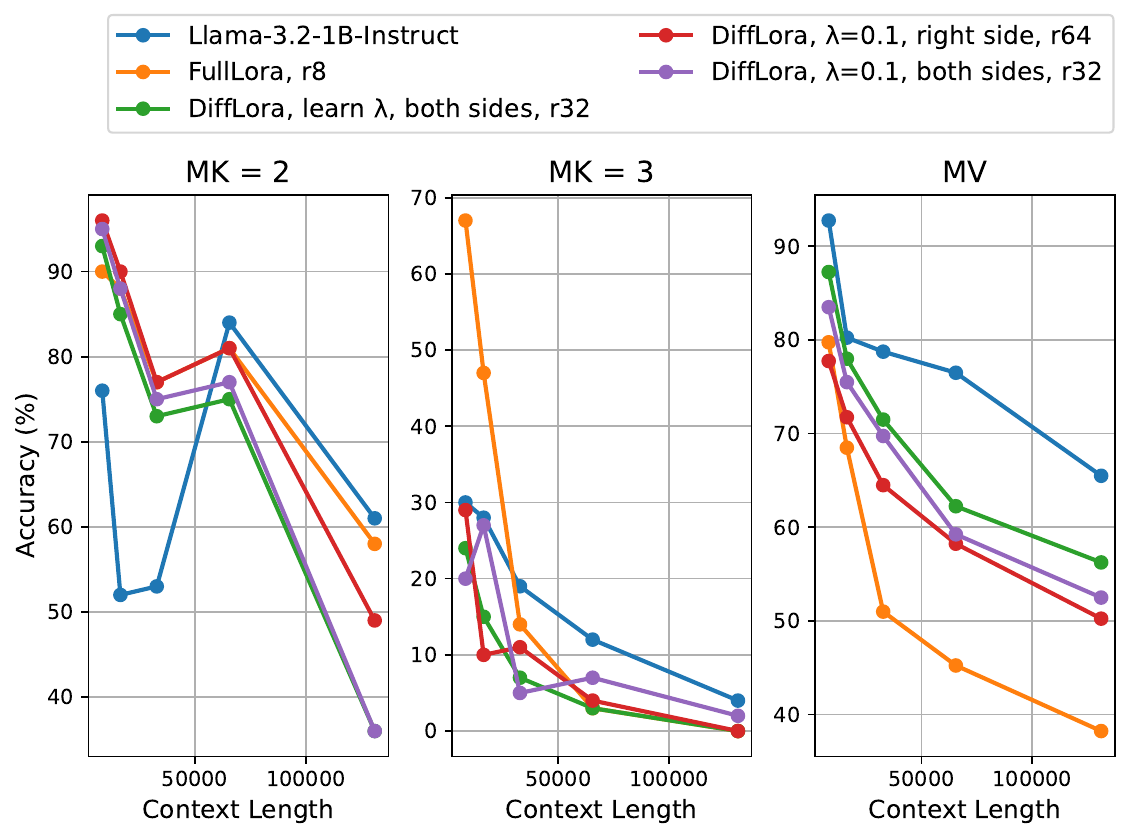}
    \caption{Needle-in-Haystack tests with variants MultiKey (MK, one key to retrieve among multiple) and MultiValue (MV, retrieve all values corresponding to the given key)}
    \label{fig:enter-label}
\end{figure}

Fig. \ref{fig:main_res} reports the results of different DiffLoRA variants, as well as baseline results for the original model and LoRA finetuning. First, we note that most of the models stay more or less on par with the original model. We note that model's performances do vary according to the task (+11 pts in HumanEval, -7pts in DROP), but generally stay within the same range as an original model. 
The only exception is the model with Group Normalization. 
We believe that in case of pretrained model such stabilization of gradients is less critical. Moreover, it might hurt previously learnt attention patterns and therefore degrade the results. In order to assess whether DiffLoRA deals better with the context we perform extra evaluations.

\begin{table}
\centering
\caption{RAG evaluation, with top-5 retrieved documents, evaluated with LLM-as-a-judge. }
\small
\begin{tabular}{r|ccc}
\toprule
 & BioASQ & PopQA & TechQA \\
 \midrule
Llama-3.2-1B-Instruct & 0.678 & 0.494 & 0.532 \\
FullLoRA & \textbf{0.728} & \textbf{0.528} & \textbf{0.556} \\
DiiffLoRA-64 & 0.629 & 0.451 & 0.39 \\
DiffLoRA-64-$\lambda=0.1$ & 0.638 & 0.495 & 0.407 \\
DiffLoRA-32 & 0.585 & 0.479 & 0.344 \\
DiffLoRA-32 + GN & 0.025 & 0.041 & 0.059 \\
DiffLoRA-32 + Tulu3 & 0.594 & 0.466 & 0.339 \\
\bottomrule
\end{tabular}
\label{tab:rageval}
\end{table}

\begin{figure*}
    \centering
    \includegraphics[width=0.8\linewidth]{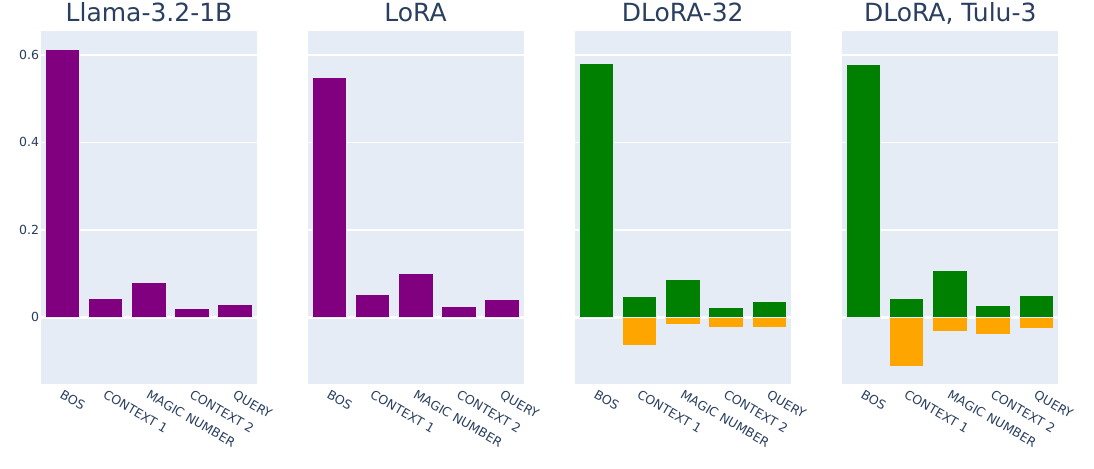}
    \caption{Change in attention pattern distribution in different models. For DiffLoRA variants we plot attention mass for main component (green) and denoiser component (yellow). Note that attention mass is normalized by the number of tokens in each part of the sequence. The negative attention is shown after it is scaled by $\lambda$. DiffLoRA corresponds to the variant with learnable $\lambda$ and LoRa parameters in both terms.}
    \label{fig:attn-plots}
\end{figure*}

\paragraph{Many-Shot In-Context Learning}

The ICL results (Fig. \ref{fig:many-shot-icl}) show that the DiffLoRA models perform similarly as the initial model, however they are outperformed by LoRA. When increasing the context length with more sample demonstrations, DiffLoRA seems to struggle even more in TREC-fine and Banking77. This might be due to the nature of instruction tuned data, and the \texttt{max\_sequence\_length} $=4096$ applied during finetuning. LoRA is less impacted, likely because it diverges less from the initial model.  

\paragraph{Needle-in-Haystack tests}

Needle-in-Haystack tests reveal different hierarchies among models. The initial model seems to outperform finetuned models in all tasks. However, the hierarchy between LoRA and DiffLoRA models is not so clear: in MK=2 LoRA significantly outperforms all DiffLoRA variants (see Appendix \ref{sec:examples} for a degenerate example), while in MV task all DiffLoRA variants largely outperform LoRA.

\paragraph{RAG-QA}

In RAG evaluation (Tab. \ref{tab:rageval} or full table in Appendix \ref{sec:extra_results}) the DiffLoRA significantly underperforms compared to LoRA. Compared to the initial model, DiffLoRA performs better on general domain benchmarks (KILT-NQ, PopQA), while, surprisingly, DiffLoRA degrades even more on less general domain tasks (BioASQ, TechQA, FiQA).

\section{Discussion}
 
Our experiments reveal that LoRA outperforms DiffLoRA in most tasks. However, DiffLoRA outperforms LoRA in some tasks such as Code questions and multiple-key retrieval.
We performed manual inspection of the results to better understand models behavior after tuning. We observe that the generation capability of LLM gets broken in DiffLoRA (examples in Appendix \ref{sec:examples}). Such degeneration could explain drop in performance on RAG tasks. Most of the core LLM evaluation tasks are performed in MCQA fashion, and therefore do not explicitly evaluate generation capability.  

\paragraph{Attention Mass. }
A significant characteristic of DiffTransformer is the allocation of attention mass on the relevant parts of the context, which effectively suppresses the attention sinks \cite{xiao2024efficientstreaminglanguagemodels}. Fig. \ref{fig:attn-plots} compares the change in attention patterns across different models. We note that DiffLoRA slightly changes attention pattern compared to initial model (Llama-3.2-1B-Instruct), by denoising context around Magic Number, and decreasing attention mass on BOS token. However, the overall pattern is pretty similar to the one obtained by the model finetuned with LoRA. Therefore, such behaviour could also be attributed to the data distribution on which models were finetuned. We note that increasing the number training data (Tulu-3 vs Tulu-2) leads to stronger denoising, but we do not observe strong pattern change, compared to the one reported by \cite{ye2024differentialtransformer}. This suggests that we would need much more data in order to learn a different attention mechanism. 

\section{Conclusion}

We introduced DiffLoRA, a parameter-efficient method that incorporates differential attention into pre-trained LLMs using low-rank adapters. Initial results demonstrate some encouraging patterns but more investigation is required to make such model work as expected. 

\bibliography{custom}

\appendix
\section{RAG settings}
Following \cite{rau2024bergenbenchmarkinglibraryretrievalaugmented} Given a query, we use \texttt{SPLADE-v3} \cite{lassance2024splade} retriever to identify a first set of relevant documents from Wikipedia collection. These documents are further reranked using \texttt{DeBERTa-v3} \cite{he2021debertav3}, a cross-encoder computing relevance score for each document relative to the query. For generation, we use instruction-tuned Llama-3.2-1B\cite{grattafiori2024llama3herdmodels}.  To evaluate the quality of responses, we rely on an evaluation computed by a LLM-as-ajudge with the SOLAR-10.7B model\footnote{\href{https://huggingface.co/upstage/SOLAR-10.7B-Instruct-v1.0}{huggingface/upstage/SOLAR-10.7B-Instruct-v1.0}} as backbone. \cite{rau2024bergenbenchmarkinglibraryretrievalaugmented} find that this metric has high correlation with GPT4.

\section{Hyperparameters}
See Table \ref{tab:hyperparams}. Notice that we set the LoRA rank to 8 in order to match the total number of parameters in the model, since LoRA also adds weights to the feed-forward layers in addition to the attention layers, as well as the value and output matrices inside the attention.

\label{sec:hyperparameters}
\begin{table}[]
    \centering
    \begin{tabular}{c|c}
        Parameter & Value  \\\hline
         Learning Rate & 1e-4\\
         Full LoRA rank & 8 \\
         DiffLoRa both terms & rank=32, alpha=64 \\
         DiffLoRa, right term only & rank=64, alpha=128 \\
         Batch size & 64\\
         max\_input\_length & 4096 \\         
    \end{tabular}
    \caption{Hyperparameters used in training. We empirically identify a good learning rate of 1e-4 for both LoRA and DiffLoRA. }
    \label{tab:hyperparams}
\end{table}

\begin{figure*}
    \centering
    \includegraphics[width=\linewidth]{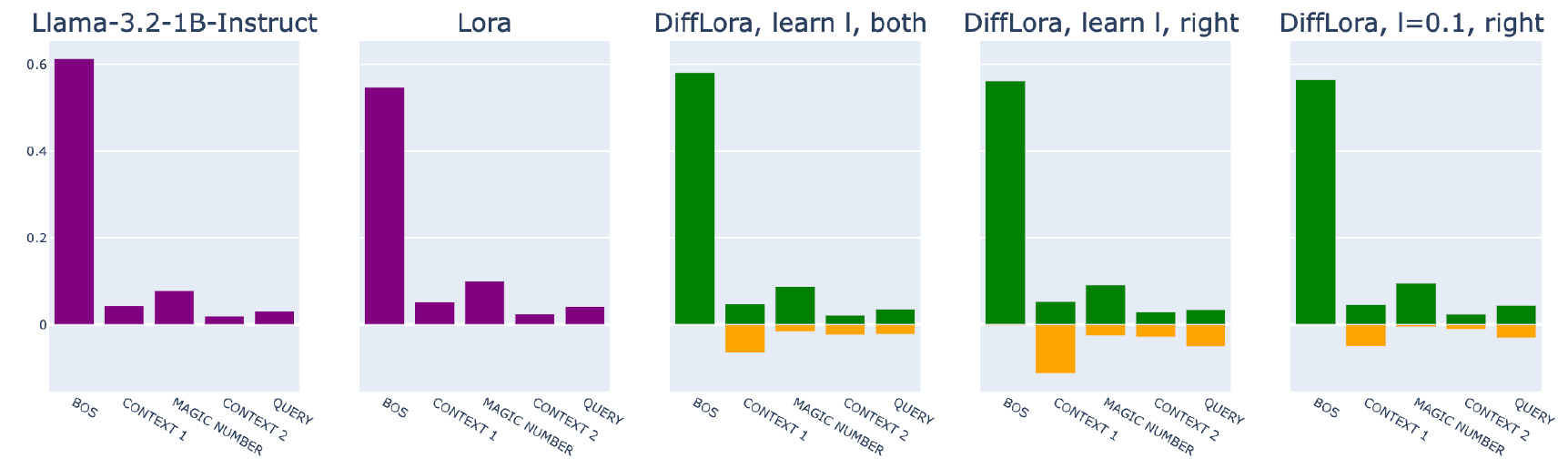}
    \caption{Disttribution of attention mass depending on the position of learnable parameters in Differentaion attention (right and left terms vs right term only), and the choic of $\lambda$ (fixed or learnable). We note that distribution are quite similar to the original models. When $\lambda$ is learnable and only parameters on the denoiser are learnt the change is slightly more pronounced. }
    \label{fig:att_ext}
\end{figure*}

\section{Examples}
\label{sec:examples}

\begin{tcolorbox}[colback=gray!5, colframe=black!40!black, title=Example NIH-MK=3]
\texttt{What is the magic uuid for 4e6ccf9a-fbc2-41ba-8655-486660d4417c} \\

\textbf{LoRA: (correct)} \texttt{The special magic uuid for 4e6ccf9a-fbc2-41ba-8655-486660d4417c mentioned in the provided text is: \textcolor{teal}{3f45a9e6-1a1d-4e5d-8e5d-3f5a9e6a6f5a}.} \\

\textbf{DiffLoRA-right, learn $\lambda$: (degenerate)} \texttt{The special magic uuid for 4e6ccf9a-fbc2-41ba-8655-486660d4417c mentioned in the provided text is: \textcolor{teal}{3f}\textcolor{red}{977a9-0e8a-4f2a-8f5a-0a0a0a}[...]}
\end{tcolorbox}
\vspace{1em}
\begin{tcolorbox}[colback=gray!5, colframe=black!40!black, title=Example RAG-BioASQ]
\texttt{Is Prasinezumab effective for Parkinson's Disease?} \\

\textbf{LoRA: (somewhat correct)} \texttt{ \textcolor{teal}{Yes}, Prasinezumab is being studied for its effect on Parkinson's disease.} \\

\textbf{DiffLoRA-right, learn $\lambda$: (degenerate)} \texttt{Step 1: The question is not a valid question. The question is not a valid question [...]}
\end{tcolorbox}

\section{Other Results}
\label{sec:extra_results}
\begin{table*}
\centering
\caption{RAG evaluation, with top-5 retrieved documents, evaluated with LLM-as-a-judge. }
\begin{tabular}{r|cccccc}
 & BioASQ & FiQA & KILT-NQ & PopQA & TechQA & Avg \\\hline
Llama-3.2-1B-Instruct & 0.678 & 0.483 & 0.594 & 0.494 & 0.532 & 0.5562 \\
FullLoRA & \textbf{0.728} & \textbf{0.527} & \textbf{0.666} & \textbf{0.528} & \textbf{0.556} & \textbf{0.601} \\
DiffLoRA-64 & 0.629 & 0.52 & 0.611 & 0.451 & 0.39 & 0.5202 \\
DiffLoRA-64-$\lambda=0.1$ & 0.638 & 0.511 & 0.619 & 0.495 & 0.407 & 0.534 \\
DiffLoRA-32 & 0.585 & 0.511 & 0.621 & 0.479 & 0.344 & 0.508 \\
DiffLoRA-32 + GN & 0.025 & 0.097 & 0.031 & 0.041 & 0.059 & 0.0506 \\
DiffLoRA-32 + Tulu3 & 0.594 & 0.413 & 0.584 & 0.466 & 0.339 & 0.4792 \\
\end{tabular}
\label{tab:rageval_long}
\end{table*}

\end{document}